\documentclass{Interspeech2024}

\interspeechcameraready

\title{Investigating Decoder-only Large Language Models for Speech-to-text Translation}

\usepackage{adjustbox}
\usepackage{multirow}

\name[affiliation={1,*}]{Chao-Wei}{Huang}
\name[affiliation={2,*}]{Hui}{Lu}
\name[affiliation={3}]{Hongyu}{Gong}
\name[affiliation={3}]{Hirofumi}{Inaguma}
\name[affiliation={3}]{Ilia}{Kulikov}
\name[affiliation={3}]{Ruslan}{Mavlyutov}
\name[affiliation={3}]{Sravya}{Popuri}

\address{
  $^1$National Taiwan University, \\
  $^2$The Chinese University of Hong Kong,
  $^3$AI at Meta}
\email{f07922069@csie.ntu.edu.tw}


\makeatletter
\def\blfootnote{\xdef\@thefnmark{}\@footnotetext}
\makeatother

\begin{document}

\maketitle

\begin{abstract}
Large language models (LLMs), known for their exceptional reasoning capabilities, generalizability, and fluency across diverse domains, present a promising avenue for enhancing speech-related tasks.
In this paper, we focus on integrating decoder-only LLMs to the task of speech-to-text translation (S2TT).
We propose a decoder-only architecture that enables the LLM to directly consume the encoded speech representation and generate the text translation.
Additionally, we investigate the effects of different parameter-efficient fine-tuning techniques and task formulation.
Our model achieves state-of-the-art performance on CoVoST 2 and FLEURS among models trained without proprietary data.
We also conduct analyses to validate the design choices of our proposed model and bring insights to the integration of LLMs to S2TT.\blfootnote{$^*$Work done during internship at Meta AI}
\end{abstract}

\section{Introduction}
The task of speech-to-text translation (S2TT) involves converting audio signals in one language into text in another, which is crucial for enabling cross-lingual communication.
Traditionally, S2TT has employed a cascaded architecture with separate automatic speech recognition (ASR) and machine translation (MT) components~\cite{ney1999speech}.
Recently, the emerging end-to-end (E2E) approach, which integrates audio encoding and text decoding into a single process, has gained popularity for the benefits of error propagation mitigation and latency reduction~\cite{inaguma2019multilingual,li-etal-2021-multilingual}.
While it has achieved significant performance improvement, S2TT still suffers from poor out-of-domain generalization and failure to capture nuanced details, e.g., slangs and cultural differences~\cite{conneau2023fleurs}. 

Large language models (LLMs) have emerged as powerful techniques for natural language processing (NLP) due to their excellent reasoning capabilities and generalizability.
They excel at generating text for a wide range of tasks based on large-scale pre-training~\cite{devlin-etal-2019-bert,brown2020language}, instruction fine-tuning~\cite{wei2021finetuned}, and reinforcement learning from human feedback~\cite{ouyang2022training,touvron2023llama2}.
LLMs are also known for their fluency and diverse domain coverage, which could potentially mitigate the generalization gap for S2TT models.
However, it is still under-explored as to how LLMs should be integrated to improve S2TT performance.

In this paper, we aim to examine various aspects of adapting decoder-only LLMs to S2TT, including architectural design, parameter-efficient fine-tuning, and taks formulations.
We propose a decoder-only architecture that directly consumes continuous speech representation instead of discretized tokens.
Our proposed model achieves state-of-the-art S2TT performance without relying on large amount of proprietary data.
Furthermore, we analyze design choices of each aspect of our experimental pipeline.
Our contribution can be summarized as the following:
\begin{itemize}
    \item We propose a decoder-only architecture for integrating LLMs to S2TT.
    \item Our proposed model outperforms state-of-the-art S2TT models on CoVoST 2 and FLEURS without training on proprietary data.
    \item We conduct analyses to validate our design choices, which we hope could facilitate future research on S2TT with LLMs.
\end{itemize}

\begin{figure*}[ht]
    \centering
    \includegraphics[width=0.8\textwidth]{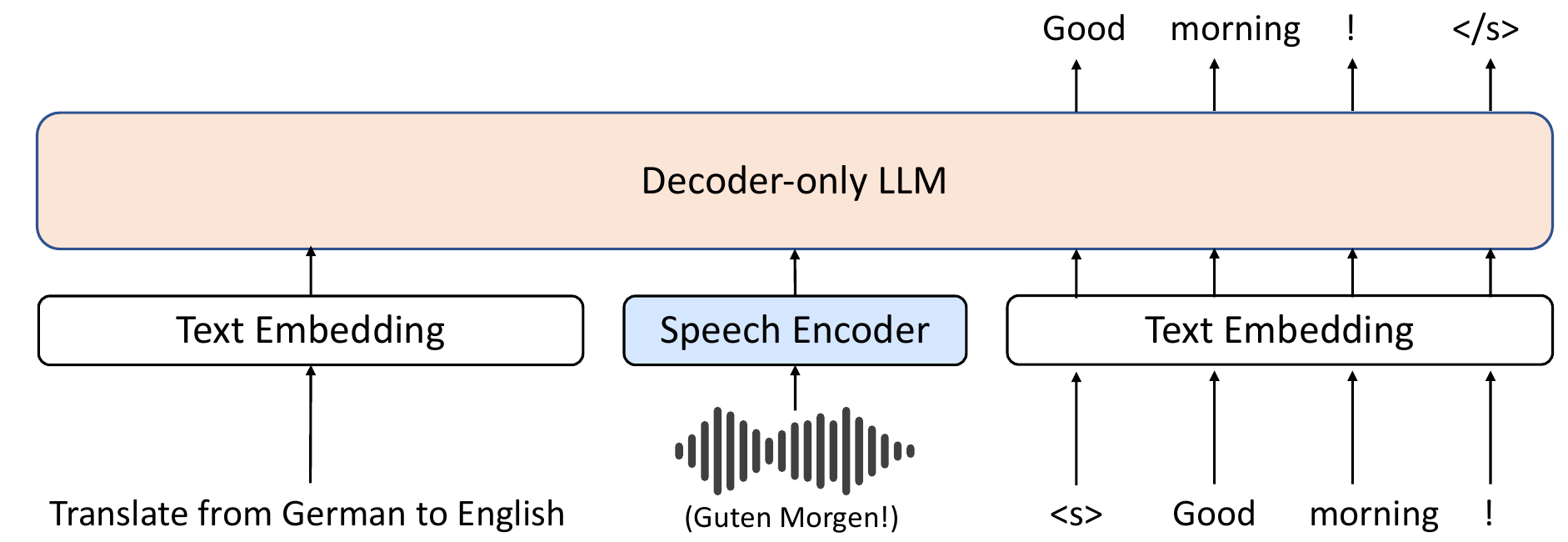}
    \caption{Illustration of our proposed decoder-only architecture.}
    \label{fig:arch}
\end{figure*}

\section{Related Work}
\subsection{Speech-to-text Translation}
Speech-to-text translation has seen significant progress, especially for end-to-end models.
To solve the data scarcity issue of training end-to-end models, multiple large-scale datasets have been collected, e.g., MuST-C~\cite{di-gangi-etal-2019-must}, CoVoST~\cite{wang2020covost}, Common Voice~\cite{ardila-etal-2020-common}, and VoxPopuli~\cite{wang-etal-2021-voxpopuli}.
Recent studies have started to focus on multilingual S2TT, where a single end-to-end model supports multiple translation directions~\cite{inaguma2019multilingual}.
The advent of pretrained models in language~\cite{devlin-etal-2019-bert,brown2020language} and speech~\cite{baevski2020wav2vec,chung2021w2v} have facilitated new state-of-the-art models that leveraged the pretrain-then-finetune paradigm~\cite{li-etal-2021-multilingual,babu22_interspeech}.

Our paper studies the integration of decoder-only LLMs to S2TT, which is still under-explored due to their new architecture and emerging capabilities.

\subsection{Speech and Audio LLMs}
With the emergence of large language models, studies have explored applying them to different modalities.
LTU~\cite{gong2024listen} fine-tuned LLMs on diverse audio datasets, thus enabling LLMs to reason given audio inputs.
Furthermore, various works have explored extending the instruction-following capability of LLMs to speech and audio inputs~\cite{wang2023slm,chu2023qwen,tang2023salmonn}.
While these methods make it possible for LLMs to handle a variety of speech and audio tasks, their performance on individual tasks often falls short of that achieved by specialized models.

Another line of research focuses on adapting LLMs to a specific speech or audio task.
Recent works have examined the integration of LLMs to automatic speech recognition, demonstrating their potential in understanding the content of speech~\cite{fathullah2023prompting,yu2023connecting}.
Similar to our work, AudioPaLM~\cite{rubenstein2023audiopalm}, Speech-LLaMA~\cite{wu2023decoder}, and SALM~\cite{chen2023salm} aimed at leveraging LLMs to improve the state-of-the-art S2TT performance.
AudioPaLM proposed to adapt LLMs to speech by discretizing speech representations and treat the discrete tokens as additional text tokens.
Such method has two drawbacks, as shown in the original paper: 1) its performance is highly dependent on the quality of the speech encoder, and 2) the discretization makes fine-tuning the speech encoder hard, which requires fine-tuning the speech encoder with ASR first~\cite{rubenstein2023audiopalm}.
Our paper demonstrates that using continuous speech representations mitigates these issues, achieving better performance while being simpler.
Speech-LLaMA and SALM both proposed briding LLMs and speech encoders with a modality adaptor and fine-tunes LLMs via LoRA~\cite{hu2022lora}.
Additionally, Speech-LLaMA introduced CTC compressor to shorten the speech input.
Our paper adopts a simpler length adaptor in our architecture, and applies LNA fine-tuning~\cite{li-etal-2021-multilingual} and demonstrates that it outperforms LoRA significantly.

\section{Our Method}
In this section, we introduce the task formulations (\S\ref{sec:formulation}), the architectural designs of our model (\S\ref{sec:arch}), how the model is trained (\S\ref{sec:training}), and parameter-efficient fine-tuning techniques (\S\ref{sec:peft}).

\subsection{Task Formulations}
\label{sec:formulation}
The task of speech-to-text translation is to translate the source speech input $S$ into the corresponding target translation $Y = \{ y_1, \cdots, y_M \}$ which is in the target language.
Following prior work~\cite{rubenstein2023audiopalm}, we define two formulations of our S2TT model:
1) the standard formulation where the model generates the target sequence directly $f\colon S \rightarrow Y$, and
2) the \textit{chained} formulation where the model first generates the transcription in the source language then the translation in the target language $f_{\text{chain}}\colon S \rightarrow \{ Y_{\text{ASR}}, Y \}$, where $Y_{\text{ASR}}$ denotes the transcription of the source speech.
It is also common to include ASR during training as an auxiliary task, which is formulated as $f_{\text{ASR}}\colon S \rightarrow Y_{\text{ASR}}$.
Therefore, we include $f$, $f_{\text{chain}}$, and $f_{\text{ASR}}$ during training for multi-task training, and perform either $f$ or $f_{\text{chain}}$ during inference.

\subsection{Architecture}
\label{sec:arch}
Our model consists of a speech encoder and a text decoder, both using the Transformer architecture~\cite{vaswani2017attention}.
An illustration of the overall architecture is shown in Figure~\ref{fig:arch}.

Our speech encoder is based on W2v-BERT~\cite{chung2021w2v}, a self-supervised pre-trained speech encoder.
For a given speech input $S$, we first convert the speech signal to fbank features with 80 mel banks, a context window of 25 ms, and a stride of 10 ms.
The speech encoder $E_{s}$ encodes the fbank features $F = \{F_1, \cdots, F_n\}$ to their corresponding hidden representations $E_{s} (F)$, where $n$ denotes the sequence length of the fbank features.
Speech frames are typically much more granular than text tokens.
Therefore, we employ a length adapter on top of the speech encoder to reduce the length of the speech representations.
The length adapter consists of a single 1-dimensional convolutional layer with a filter size and stride of $k$, which reduces the length of the speech representations by $k$-fold.

The text decoder is based on LLaMA-2~\cite{touvron2023llama2}, a decoder-only large language model pre-trained on 2 trillion text tokens with a language modeling objective.
The speech inputs and text inputs are encoded with their corresponding encoders, i.e., speech encoder for speech inputs and text embedding layer for text inputs.
Subsequently, the encoded representations are concatenated and fed to the transformer decoder.
In other words, we treat the encoded speech representations $S$ the same as the text embeddings, without discretizing them as done in prior work~\cite{rubenstein2023audiopalm}.
A triangular mask is appied to the self-attention layers to restrict tokens from atteding to latter positions.
More formally, given an interleaving sequence of text and speech sequences $X = \{X^1, F, X^2 \}$, where $X^i = \{ x^i_i, \cdots, x^i_{|x^i|} \}$ denotes a text sequence, $X^1$ denotes the prefix text, and $X^2$ denotes the suffix text.
After encoding, the input sequence to the transformer decoder will be $ \mathbf{X} = \{ \text{Emb}(X^1), E_s(F), \text{Emb}(X^2) \}$, where $\text{Emb}$ denotes the text embedding layer.
Note that we flatten the sequences in $\mathbf{X}$ before processing them with the decoder.
Finally, we apply a linear transformation to the decoder outputs to obtain the logits for predicting the next token $\mathbf{O} = W^\top D(\mathbf{X})$, where $D$ denotes the transformer decoder and $W \in \mathbb{R}^{h \times |V|}$ is a trainable matrix where $|V|$ denotes the vocabulary size.

\begin{table*}[ht]
\begin{adjustbox}{width=\textwidth}
\setlength\tabcolsep{3pt}
\begin{tabular}{lcccccccccccccccccccccc}
\toprule
    & Ar   & Ca   & Cy   & De   & Es   & Et   & Fa   & Fr   & Id   & It   & Ja   & Lv   & Mn   & Nl   & Pt   & Ru   & Sl   & Sv   & Ta   & Tr  & Zh   & \bf Avg  \\
\midrule
\multicolumn{23}{l}{\bf Trained with Proprietary Data} \\
\midrule
Whisper-large~\cite{radford2023robust} & 39.7 & 31.8 & 18.0   & 21.5 & 36.3 & 15.0   & 36.4 & 48.1 & 30.9 & 26.1 & 0.1  & 13.9 & 41.2 & 19.3 & 51.6 & 43.3 & 21.6 & 40.1 & 42.9 & 4.2 & 28.3 & 29.1 \\
USM-M~\cite{zhang2023google}         & -    & -    & -    & -    & -    & -    & -    & -    & -    & -    & -    & -    & -    & -    & -    & -    & -    & -    & -    & -   & -    & 30.7 \\
Speech-LLaMA~\cite{wu2023decoder}  & 28.2 & -    & -    & 27.1 & 27.9 & 18.7 & -    & 25.2 & -    & 25.9 & 19.9 & -    & -    & 36.5 & 32.0   & 36.8 & 22.7 & 29.0   & -    & -   & 12.3 & -    \\
AudioPaLM~\cite{rubenstein2023audiopalm}   & 48.7 & 38.4 & 25.5 & 13.7 & 43.4 & 30.0   & 44.8 & 56.2 & 44.3 & 25.9 & 7.6  & 35.0   & 48.3 & 29.4 & 57.3 & 55.6 & 42.6 & 44.2 & 53.3 & 9.0   & 41.0   & 37.8 \\
\midrule
\multicolumn{23}{l}{\bf Trained with Public Data Only}  \\
\midrule 
XLS-R~\cite{babu22_interspeech}         & 17.1 & 33.8 & 9.4  & 14.0   & 33.6 & 11.1 & 37.6 & 16.5 & 34.9 & 3.5  & 1.6  & 19.5 & 31.7 & 12.9 & 41.8 & 39.5 & 19.6 & 39.2 & 29.6 & 0.5 & 16.7 & 22.1 \\
ComSL-large~\cite{le2024comsl}     & -    & -    & -    & -    & -    & -    & -    & -    & -    & -    & -    & -    & -    & -    & -    & -    & -    & -    & -    & -   & -    & 31.5 \\
AudioPaLM\textsuperscript{\textdagger}~\cite{rubenstein2023audiopalm}     & -    & -    & -    & -    & -    & -    & -    & -    & -    & -    & -    & -    & -    & -    & -    & -    & -    & -    & -    & -   & -    & 33.1 \\
W2vBERT+NLLB      & 42.0 & 38.4 & 18.3 & 52.0 & 39.3 & \bf 23.6 & 41.3 & 47.3 & 39.4 & 18.1 & \bf 3.5  & 18.4 & 43.0 & \bf 27.2 & 50.8 & 51.7 & 36.9 & 42.2 & 40.6 & \bf 6.2 & 33.2 & 34.0 \\
Ours          & \bf 45.8 & \bf 39.5 & \bf 22.4 & \bf 56.9 & \bf 41.2 & 20.4 & \bf 44.5 & \bf 54.5 & \bf 42.9 & \bf 24.4 & 0.9  & \bf 21.9 & \bf 46.8 & 26.3 & \bf 56.1 & \bf 53.3 & \bf 42.7 & \bf 45.1 & \bf 53.7 & 5.3 & \bf 34.4 & \bf 37.1 \\
\bottomrule
\end{tabular}
\end{adjustbox}
\vspace{2pt}
\caption{Main results on the X-En test sets of CoVoST 2 (\%). We report corpus BLEU scores computed with SacreBLEU. The best results among models trained with public data are bolded. \textsuperscript{\textdagger}The result reported in the AudioPaLM paper~\cite{rubenstein2023audiopalm} when trained on only public datasets.}
\label{tab:main}
\end{table*}

\subsection{Training}
\label{sec:training}
As described above, we include three formulations, i.e., $f$, $f_{\text{chain}}$, and $f_{\text{ASR}}$, for multi-task training.
To let our model distinguish among tasks, we provide different instructions in natural language for each task $t$.
The instructions include a description of the task, the source language, and the target language.
We format the instruction $I$ and the source speech $S$ into the input sequence $X$ with a template.
The target sequence for training is formatted as:
\begin{equation*}
    Y' = 
        \begin{cases}
        \text{Translation: } Y & \text{if } t = f \\
        \text{Transcription: } Y_{\text{ASR}} & \text{if } t = f_{\text{ASR}} \\
        \text{Transcription: } Y_{\text{ASR}} \text{ Translation: } Y & \text{if } t = f_{\text{chain}}.
        \end{cases}
\end{equation*}

Given a source speech $S$, an instruction $I$, and the formatted target sequence $Y'$, the training objective is to minimize the S2TT loss:
\begin{equation*}
    \mathcal{L}(S, Y') = - \frac{1}{M'} \sum_{i=1}^{M'} \text{log} P(y'_i \mid S, I, Y'_{<i})
\end{equation*}
where $M'$ denotes the length of $Y'$ and $P(y'_i \mid S, I, Y'_{<i})$ denotes the probability of $y'_i$ predicted by the model given the source speech and the prior tokens $Y'_{<i}$ in the target sequence.
The predicted probability is obtained by applying the softmax function to the logits $\mathbf{O}$.

\subsection{Parameter-efficient Fine-tuning}
\label{sec:peft}
Large language models have billions of parameters, making it computationally expensive and inefficient to fine-tune all of the parameters during training.
It is common to apply parameter-efficient fine-tuning techniques when fine-tuning LLMs on downstream tasks to improve efficiency and mitigate catastrophic forgetting.
To this end, we employ and compare two parameter-efficient fine-tuning techniques in this paper: LNA fine-tuning~\cite{li-etal-2021-multilingual} and Low Rank Adaptation (LoRA)~\cite{hu2022lora}.

\subsubsection{LNA Fine-tuning}
LayerNorm and Attention (LNA) fine-tuning adapts pretrained language and speech models to S2TT by fine-tuning only the layer normalization and the multi-head attention layers~\cite{li-etal-2021-multilingual}.
This method greatly reduces the number of trainable parameters during fine-tuning and avoids catastrophic forgetting, thus improving the downstream performance for multilingual speech-to-text translation.
Since the pretrained language model we use is a decoder-only transformer model, we apply LNA fine-tuning and fine-tune only the layer normalization and the self-attention layers in the transformer decoder.

\subsubsection{Low Rank Adaptation (LoRA)}
LoRA injects trainable rank decomposition matrices into the projections layers of a transformer model, which serves as a residual path in addition to a projection layer.
During fine-tuning, only the decomposition matrices are updated, while all of the pretrained parameters are frozen.
Thus, the number of trainable parameters is significantly reduced.
The decomposition matrices can be merged into the original projection matrix after fine-tuning.
Therefore, there is no additional computation nor additional parameters compared to the pretrained transformer model during inference, making LoRA a common technique for adapting large language models efficiently.

\section{Experiments}

\subsection{Experimental Setup}
We train and evaluate our models on publicly available datasets.
For training, we use CoVoST2~\cite{wang2020covost}, Common Voice 11~\cite{ardila-etal-2020-common}, and VoxPopuli~\cite{wang-etal-2021-voxpopuli} datasets.
CoVoST-2 is a speech-to-text translation dataset consisting of 21 languages.
The dataset includes human-labeled translation pairs from 21 languages to English (X-En), and from English to 15 languages (En-X).
Common Voice is a collection of speech-text pairs where the speech was recorded by annotators given the text transcription.
VoxPopuli consists of speech from the European Parliament with the corresponding transcriptions and interpretations in 15 languages.

We conduct in-domain evaluation on the test sets of CoVoST 2.
Additionally, we perform zero-shot evaluation on FLEURS~\cite{conneau2023fleurs}, a dataset that aims to evaluate the out-of-domain generalizability of speech translation models.
Note that for all datasets, we only use the directions that are present in CoVoST2.
We report BLEU scores from SacreBLEU and additionally the model-based COMET score with the model \textit{wmt22-comet-da}~\cite{rei-etal-2022-comet}.

\subsection{Implementation Details}
We employ a pretrained W2v-BERT~\cite{chung2021w2v} model that was released in~\cite{barrault2023seamlessm4t} with 600M parameters that is pretrained on 4 million hours of speech data with a self-supervised objective as the speech encoder.
The text decoder is initialized with \textit{LLaMA2-7B-chat}~\cite{touvron2023llama2}.

We implement our models, training, and evaluation procedures with the Fairseq2 library\footnote{https://github.com/facebookresearch/fairseq2}.
During training, the effective batch size is set to 800K speech frames, or 8000 seconds of speech inputs.
We optimize the model with the AdamW optimizer and set the learning rate to 1e-4.
The learning rate is warmed up for 5000 steps and linearly decayed until the maximum number of steps is reached, which is set to 60000.
We fine-tune all parameters of the speech encoder and apply parameter-efficient fine-tuning methods to the text decoder.
All experiments are conducted on 32 NVIDIA A100 GPUs.

\begin{table}[ht]
\begin{adjustbox}{width=\linewidth}
\setlength\tabcolsep{5pt}
\begin{tabular}{l|cc|cc}
\toprule
 & \multicolumn{2}{c|}{CoVoST 2} & \multicolumn{2}{c}{FLEURS} \\
 & \multicolumn{1}{c}{BLEU} & \multicolumn{1}{c|}{COMET} & \multicolumn{1}{c}{BLEU} & \multicolumn{1}{c}{COMET} \\
\midrule
\multicolumn{5}{l}{\bf Encoder-Decoder} \\
\midrule
W2vBERT+NLLB   & 33.8 & 80.1 & 22.7 & 76.4 \\
W2vBERT+LLaMA2 & 33.3 & 79.7 & 18.2 & 74.2 \\
\midrule
\multicolumn{5}{l}{\bf Decoder-only} \\
\midrule
Ours & \bf 36.3 & \bf 81.4 & \bf 23.4 & \bf 77.6 \\
\bottomrule
\end{tabular}
\end{adjustbox}
\vspace{2pt}
\caption{Results of different architectures (\%). We report the average BLEU score and COMET score on the 21 X-En directions on CoVoST 2 and FLEURS.}
\label{tab:arch}
\end{table}

\begin{table}[ht]
\begin{adjustbox}{width=\linewidth}
\begin{tabular}{ccccc}
\toprule
\multicolumn{1}{l}{}  & Rank & Layers               & CoVoST 2 & FLEURS \\
\midrule
Freeze W2vBERT & - & - & 15.6 & 3.8 \\
Freeze decoder                & -    & -                    & 29.9     & 19.0     \\
\midrule
\multirow{3}{*}{LoRA} & 8    & q, v          & 32.1     & 20.8   \\
                      & 32   & q, k, v, o & 32.7     & 21.1   \\
                      & 32   & All linear    & 33.0       & 21.4   \\
\midrule
LNA                   & -    & -                    & \bf 36.3     & \bf 23.4  \\
\bottomrule
\end{tabular}
\end{adjustbox}
\vspace{2pt}
\caption{Results of different parameter-efficient fine-tuning methods (\%). Rank and Layers refer to the configuration of LoRA. The notations q, k, v, o denote the query, key, value, output layers of the self-attention layers respectively.}
\label{tab:peft}
\end{table}

\subsection{Baseline Methods}
We compare our model with various state-of-the-art baselines that were trained on the same set of public datasets as our method, i.e., CoVoST 2, Common Voice, and VoxPopuli.
XLS-R~\cite{babu22_interspeech} is a self-supervised cross-lingual speech representation model.
ComSL~\cite{le2024comsl} conducts self-training on the Common Voice dataset.
Additionally, we implement an encoder-decoder baseline with W2vBERT as the speech encoder and NLLB~\cite{costa2022no} 1.3B as the text decoder.

We also compare our model with models trained with proprietary data.
Whisper~\cite{radford2023robust} trains a robust speech recognition and translation model with large amounts of weak supervisions.
USM~\cite{zhang2023google} is an universal speech model pretrained with 12 million hours of speech data.
Speech-LLaMA~\cite{wu2023decoder} shares a similar architecture with our model and was trained with in-house data and LoRA~\cite{hu2022lora}.
AudioPaLM~\cite{rubenstein2023audiopalm} is the state-of-the-art method on CoVoST2 which is trained on proprietary data.
We also include a variant of AudioPaLM that is trained on public datasets only for a fair comparison, which is reported in the paper~\cite{rubenstein2023audiopalm}.

\subsection{Results}
The main results on CoVoST 2 are reported in Table~\ref{tab:main}.
Our model achieves an average BLEU score of 37.1, which is the new state-of-the-are performance among models trained with public data only.
Notably, our model outperforms the AudioPaLM variant which was trained on only public datasets, demonstrating the superiority of our proposed method.
When compared to models trained with proprietary data, our model outperforms all of them and achieves comparable performance to AudioPaLM.
These results demonstrate that our method integrates LLMs to S2TT efficiently and effectively.

\begin{table}[ht]
\centering
\setlength\tabcolsep{6pt}
\begin{tabular}{l|cc}
\toprule
 & CoVoST 2 & FLEURS \\
\midrule
Ours                 & \bf 37.1                         & \bf 23.4                       \\
\quad - $f_{\text{chain}}$           & 35.4                         & 22.4                       \\
\quad - $f_{\text{ASR}}$             & 36.4                         & 22.7                       \\
\quad - $f_{\text{ASR}} \mathbin{\&} f_{\text{chain}}$ & 35.8                         & 22.5 \\
\bottomrule
\end{tabular}
\vspace{2pt}
\caption{Results of formulation ablation (\%).}
\label{tab:ablation}
\end{table}

\section{Discussion}
In this section, we conduct various experiments to analyze and discuss the details of our proposed method.

\subsection{Architectural Design}
With decoder-only LLMs, it is unclear as to which architecture performs the best for S2TT.
We compare our decoder-only architecture with encoder-decoder models, with NLLB~\cite{costa2022no} and LLaMA-2~\cite{touvron2023llama2} as the text decoder.
As shown in Table~\ref{tab:arch}, our model significantly outperforms the encoder-decoder counterpart on both CoVoST 2 and FLEURS.
Furthermore, encoder-decoder with LLaMA 2 even underperforms NLLB, demonstrating that encoder-decoder architecture are unsuitable for decoder-only LLMs.
We hypothesize that it is the newly introduced encoder-decoder attention layers which are not pretrained that degrade the performance of encoder-decoder models.

\subsection{Parameter-efficient Fine-tuning}
We compare LNA fine-tuning, LoRA, and the effect of freezing pretrained models.
As shown in Table~\ref{tab:peft}, LNA fine-tuning significantly outperforms LoRA with various configurations.
This result suggests that adopting LoRA, as done in prior work such as Speech-LLaMA~\cite{wu2023decoder}, is suboptimal for S2TT.
Freezing the text decoder during fine-tuning yields even worse performance than LoRA, demonstrating the importance of fine-tuning the text decoder.
Finally, freezing the speech encoder results in detrimental performance degradation.
This result shows that fine-tuning the speech encoder is crucial for aligning the speech representation with the text inputs.
We hypothesize that this leads to the underperformance of AudioPaLM with encoders that are not fine-tuned with ASR~\cite{rubenstein2023audiopalm}, since the discretization of speech representations makes fine-tuning the speech encoder non-trivial.

\subsection{Ablation of Formulations}
Table~\ref{tab:ablation} shows the results of various combination of the formulations.
Removing either $f_{\text{ASR}}$ or $f_{\text{chain}}$ degrades the S2TT performance.
Notably, training with $f$ and $f_{\text{ASR}}$ slightly underperforms $f$, showing that multi-task training with ASR does not always improve performance.

\section{Conclusion}
In this paper, we propose a decoder-only architecture that adapts a decoder-only LLM to the speech-to-text translation task.
Our proposed method is simple and effective, achieving state-of-the-art performance and is comparable to the best-performing proprietary model.
We conduct additional analyses to examine the effect of different design choices regarding architectural design, parameter-efficient fine-tuning, and task formulations.
We hope that our findings could facilitate future work on leveraging LLMs in the S2TT task.

\bibliographystyle{IEEEtran}
\bibliography{mybib}

\begin{thebibliography}{10}
\providecommand{\url}[1]{#1}
\csname url@samestyle\endcsname
\providecommand{\newblock}{\relax}
\providecommand{\bibinfo}[2]{#2}
\providecommand{\BIBentrySTDinterwordspacing}{\spaceskip=0pt\relax}
\providecommand{\BIBentryALTinterwordstretchfactor}{4}
\providecommand{\BIBentryALTinterwordspacing}{\spaceskip=\fontdimen2\font plus
\BIBentryALTinterwordstretchfactor\fontdimen3\font minus \fontdimen4\font\relax}
\providecommand{\BIBforeignlanguage}[2]{{%
\expandafter\ifx\csname l@#1\endcsname\relax
\typeout{** WARNING: IEEEtran.bst: No hyphenation pattern has been}%
\typeout{** loaded for the language `#1'. Using the pattern for}%
\typeout{** the default language instead.}%
\else
\language=\csname l@#1\endcsname
\fi
#2}}
\providecommand{\BIBdecl}{\relax}
\BIBdecl

\bibitem{ney1999speech}
H.~Ney, ``Speech translation: Coupling of recognition and translation,'' in \emph{1999 IEEE International Conference on Acoustics, Speech, and Signal Processing. Proceedings. ICASSP99 (Cat. No. 99CH36258)}, vol.~1.\hskip 1em plus 0.5em minus 0.4em\relax IEEE, 1999, pp. 517--520.

\bibitem{inaguma2019multilingual}
H.~Inaguma, K.~Duh, T.~Kawahara, and S.~Watanabe, ``Multilingual end-to-end speech translation,'' in \emph{2019 IEEE Automatic Speech Recognition and Understanding Workshop (ASRU)}.\hskip 1em plus 0.5em minus 0.4em\relax IEEE, 2019, pp. 570--577.

\bibitem{li-etal-2021-multilingual}
X.~Li, C.~Wang, Y.~Tang, C.~Tran, Y.~Tang, J.~Pino, A.~Baevski, A.~Conneau, and M.~Auli, ``Multilingual speech translation from efficient finetuning of pretrained models,'' in \emph{Proceedings of ACL-IJCNLP 2021}, 2021, pp. 827--838.

\bibitem{conneau2023fleurs}
A.~Conneau, M.~Ma, S.~Khanuja, Y.~Zhang, V.~Axelrod, S.~Dalmia, J.~Riesa, C.~Rivera, and A.~Bapna, ``Fleurs: Few-shot learning evaluation of universal representations of speech,'' in \emph{2022 IEEE Spoken Language Technology Workshop (SLT)}.\hskip 1em plus 0.5em minus 0.4em\relax IEEE, 2023, pp. 798--805.

\bibitem{devlin-etal-2019-bert}
J.~Devlin, M.-W. Chang, K.~Lee, and K.~Toutanova, ``{BERT}: Pre-training of deep bidirectional transformers for language understanding,'' in \emph{Proceedings NAACL-HLT 2019}, 2019, pp. 4171--4186.

\bibitem{brown2020language}
T.~Brown, B.~Mann, N.~Ryder, M.~Subbiah, J.~D. Kaplan, P.~Dhariwal, A.~Neelakantan, P.~Shyam, G.~Sastry, A.~Askell \emph{et~al.}, ``Language models are few-shot learners,'' \emph{Advances in neural information processing systems}, vol.~33, pp. 1877--1901, 2020.

\bibitem{wei2021finetuned}
J.~Wei, M.~Bosma, V.~Y. Zhao, K.~Guu, A.~W. Yu, B.~Lester, N.~Du, A.~M. Dai, and Q.~V. Le, ``Finetuned language models are zero-shot learners,'' \emph{arXiv:2109.01652}, 2021.

\bibitem{ouyang2022training}
L.~Ouyang, J.~Wu, X.~Jiang, D.~Almeida, C.~Wainwright, P.~Mishkin, C.~Zhang, S.~Agarwal, K.~Slama, A.~Ray \emph{et~al.}, ``Training language models to follow instructions with human feedback,'' \emph{Advances in Neural Information Processing Systems}, vol.~35, pp. 27\,730--27\,744, 2022.

\bibitem{touvron2023llama2}
H.~Touvron, L.~Martin, K.~Stone, P.~Albert, A.~Almahairi, Y.~Babaei, N.~Bashlykov, S.~Batra, P.~Bhargava, S.~Bhosale \emph{et~al.}, ``Llama 2: Open foundation and fine-tuned chat models,'' \emph{arXiv:2307.09288}, 2023.

\bibitem{di-gangi-etal-2019-must}
M.~A. Di~Gangi, R.~Cattoni, L.~Bentivogli, M.~Negri, and M.~Turchi, ``{M}u{ST}-{C}: a {M}ultilingual {S}peech {T}ranslation {C}orpus,'' in \emph{Proceedings of NAACL-HLT 2019}, 2019, pp. 2012--2017.

\bibitem{wang2020covost}
C.~Wang, A.~Wu, and J.~Pino, ``Covost 2 and massively multilingual speech-to-text translation,'' \emph{arXiv:2007.10310}, 2020.

\bibitem{ardila-etal-2020-common}
R.~Ardila, M.~Branson, K.~Davis, M.~Kohler, J.~Meyer, M.~Henretty, R.~Morais, L.~Saunders, F.~Tyers, and G.~Weber, ``Common voice: A massively-multilingual speech corpus,'' in \emph{Proceedings of LREC 2020}, 2020, pp. 4218--4222.

\bibitem{wang-etal-2021-voxpopuli}
C.~Wang, M.~Riviere, A.~Lee, A.~Wu, C.~Talnikar, D.~Haziza, M.~Williamson, J.~Pino, and E.~Dupoux, ``{V}ox{P}opuli: A large-scale multilingual speech corpus for representation learning, semi-supervised learning and interpretation,'' in \emph{Proceedings of ACL-IJCNLP 2021}, 2021, pp. 993--1003.

\bibitem{baevski2020wav2vec}
A.~Baevski, Y.~Zhou, A.~Mohamed, and M.~Auli, ``wav2vec 2.0: A framework for self-supervised learning of speech representations,'' \emph{Advances in neural information processing systems}, vol.~33, pp. 12\,449--12\,460, 2020.

\bibitem{chung2021w2v}
Y.-A. Chung, Y.~Zhang, W.~Han, C.-C. Chiu, J.~Qin, R.~Pang, and Y.~Wu, ``W2v-bert: Combining contrastive learning and masked language modeling for self-supervised speech pre-training,'' in \emph{2021 IEEE Automatic Speech Recognition and Understanding Workshop (ASRU)}.\hskip 1em plus 0.5em minus 0.4em\relax IEEE, 2021, pp. 244--250.

\bibitem{babu22_interspeech}
A.~Babu, C.~Wang, A.~Tjandra, K.~Lakhotia, Q.~Xu, N.~Goyal, K.~Singh, P.~{von Platen}, Y.~Saraf, J.~Pino, A.~Baevski, A.~Conneau, and M.~Auli, ``{XLS-R: Self-supervised Cross-lingual Speech Representation Learning at Scale},'' in \emph{Proc. Interspeech 2022}, 2022, pp. 2278--2282.

\bibitem{gong2024listen}
Y.~Gong, H.~Luo, A.~H. Liu, L.~Karlinsky, and J.~R. Glass, ``Listen, think, and understand,'' in \emph{The Twelfth International Conference on Learning Representations}, 2024.

\bibitem{wang2023slm}
M.~Wang, W.~Han, I.~Shafran, Z.~Wu, C.-C. Chiu, Y.~Cao, N.~Chen, Y.~Zhang, H.~Soltau, P.~K. Rubenstein \emph{et~al.}, ``Slm: Bridge the thin gap between speech and text foundation models,'' in \emph{ASRU 2023}.\hskip 1em plus 0.5em minus 0.4em\relax IEEE, 2023, pp. 1--8.

\bibitem{chu2023qwen}
Y.~Chu, J.~Xu, X.~Zhou, Q.~Yang, S.~Zhang, Z.~Yan, C.~Zhou, and J.~Zhou, ``Qwen-audio: Advancing universal audio understanding via unified large-scale audio-language models,'' \emph{arXiv:2311.07919}, 2023.

\bibitem{tang2023salmonn}
C.~Tang, W.~Yu, G.~Sun, X.~Chen, T.~Tan, W.~Li, L.~Lu, Z.~Ma, and C.~Zhang, ``Salmonn: Towards generic hearing abilities for large language models,'' \emph{arXiv:2310.13289}, 2023.

\bibitem{fathullah2023prompting}
Y.~Fathullah, C.~Wu, E.~Lakomkin, J.~Jia, Y.~Shangguan, K.~Li, J.~Guo, W.~Xiong, J.~Mahadeokar, O.~Kalinli \emph{et~al.}, ``Prompting large language models with speech recognition abilities,'' \emph{arXiv:2307.11795}, 2023.

\bibitem{yu2023connecting}
W.~Yu, C.~Tang, G.~Sun, X.~Chen, T.~Tan, W.~Li, L.~Lu, Z.~Ma, and C.~Zhang, ``Connecting speech encoder and large language model for asr,'' \emph{arXiv:2309.13963}, 2023.

\bibitem{rubenstein2023audiopalm}
P.~K. Rubenstein, C.~Asawaroengchai, D.~D. Nguyen, A.~Bapna, Z.~Borsos, F.~d.~C. Quitry, P.~Chen, D.~E. Badawy, W.~Han, E.~Kharitonov \emph{et~al.}, ``Audiopalm: A large language model that can speak and listen,'' \emph{arXiv:2306.12925}, 2023.

\bibitem{wu2023decoder}
J.~Wu, Y.~Gaur, Z.~Chen, L.~Zhou, Y.~Zhu, T.~Wang, J.~Li, S.~Liu, B.~Ren, L.~Liu \emph{et~al.}, ``On decoder-only architecture for speech-to-text and large language model integration,'' in \emph{2023 IEEE Automatic Speech Recognition and Understanding Workshop (ASRU)}.\hskip 1em plus 0.5em minus 0.4em\relax IEEE, 2023, pp. 1--8.

\bibitem{chen2023salm}
Z.~Chen, H.~Huang, A.~Andrusenko, O.~Hrinchuk, K.~C. Puvvada, J.~Li, S.~Ghosh, J.~Balam, and B.~Ginsburg, ``Salm: Speech-augmented language model with in-context learning for speech recognition and translation,'' \emph{arXiv:2310.09424}, 2023.

\bibitem{hu2022lora}
E.~J. Hu, Y.~Shen, P.~Wallis, Z.~Allen-Zhu, Y.~Li, S.~Wang, L.~Wang, and W.~Chen, ``Lo{RA}: Low-rank adaptation of large language models,'' in \emph{International Conference on Learning Representations}, 2022.

\bibitem{vaswani2017attention}
A.~Vaswani, N.~Shazeer, N.~Parmar, J.~Uszkoreit, L.~Jones, A.~N. Gomez, {\L}.~Kaiser, and I.~Polosukhin, ``Attention is all you need,'' \emph{Advances in neural information processing systems}, vol.~30, 2017.

\bibitem{radford2023robust}
A.~Radford, J.~W. Kim, T.~Xu, G.~Brockman, C.~McLeavey, and I.~Sutskever, ``Robust speech recognition via large-scale weak supervision,'' in \emph{International Conference on Machine Learning}.\hskip 1em plus 0.5em minus 0.4em\relax PMLR, 2023, pp. 28\,492--28\,518.

\bibitem{zhang2023google}
Y.~Zhang, W.~Han, J.~Qin, Y.~Wang, A.~Bapna, Z.~Chen, N.~Chen, B.~Li, V.~Axelrod, G.~Wang \emph{et~al.}, ``Google usm: Scaling automatic speech recognition beyond 100 languages,'' \emph{arXiv:2303.01037}, 2023.

\bibitem{le2024comsl}
C.~Le, Y.~Qian, L.~Zhou, S.~Liu, Y.~Qian, M.~Zeng, and X.~Huang, ``Comsl: A composite speech-language model for end-to-end speech-to-text translation,'' \emph{Advances in Neural Information Processing Systems}, vol.~36, 2024.

\bibitem{rei-etal-2022-comet}
R.~Rei, J.~G. C.~de Souza, D.~Alves, C.~Zerva, A.~C. Farinha, T.~Glushkova, A.~Lavie, L.~Coheur, and A.~F.~T. Martins, ``{COMET}-22: Unbabel-{IST} 2022 submission for the metrics shared task,'' in \emph{Proceedings of the Seventh Conference on Machine Translation (WMT)}, 2022, pp. 578--585.

\bibitem{barrault2023seamlessm4t}
L.~Barrault, Y.-A. Chung, M.~C. Meglioli, D.~Dale, N.~Dong, P.-A. Duquenne, H.~Elsahar, H.~Gong, K.~Heffernan, J.~Hoffman \emph{et~al.}, ``Seamlessm4t-massively multilingual \& multimodal machine translation,'' \emph{arXiv:2308.11596}, 2023.

\bibitem{costa2022no}
M.~R. Costa-juss{\`a}, J.~Cross, O.~{\c{C}}elebi, M.~Elbayad, K.~Heafield, K.~Heffernan, E.~Kalbassi, J.~Lam, D.~Licht, J.~Maillard \emph{et~al.}, ``No language left behind: Scaling human-centered machine translation,'' \emph{arXiv:2207.04672}, 2022.

\end{thebibliography}

\end{document}